\documentclass[a4paper]{article}

\usepackage[T1]{fontenc}
\usepackage[utf8]{inputenc}
\usepackage{times}            % Times Roman body font (ECAI style)
\usepackage[margin=2cm,columnsep=0.6cm]{geometry}
\usepackage{multicol}
\usepackage{titlesec}
\usepackage{amsmath,amssymb}
\usepackage{graphicx}
\usepackage{booktabs}
\usepackage{tabularx}
\usepackage{array}
\usepackage{multirow}
\usepackage{caption}
\usepackage{float}
\usepackage{enumitem}
\usepackage[hidelinks]{hyperref} % arXiv: clickable but subtle (black) links
\usepackage{pifont}           % for \ding checkmarks
\usepackage{url}
\usepackage{microtype}
\usepackage{xcolor}
\hypersetup{
  pdftitle={From Metaheuristics to Exact Methods: A CP-SAT Approach for
            Multi-Objective Healthcare Workforce Scheduling},
  pdfauthor={Vipul Patel, Anirudh Deodhar, Dagnachew Birru},
  pdfkeywords={constraint programming, CP-SAT, healthcare workforce scheduling,
               nurse rostering, multi-objective optimization}
}

\newcommand{\cmark}{\ding{51}}   % ✓
\newcommand{\xmark}{\ding{55}}   % ✘

\titleformat{\section}{\normalsize\bfseries}{\thesection}{0.5em}{}
\titleformat{\subsection}{\normalsize\bfseries\itshape}{\thesubsection}{0.5em}{}
\titleformat{\subsubsection}{\normalsize\itshape}{\thesubsubsection}{0.5em}{}
\titlespacing*{\section}{0pt}{10pt}{4pt}
\titlespacing*{\subsection}{0pt}{8pt}{3pt}

\setlist{nosep,leftmargin=1.5em}

\newcolumntype{L}[1]{>{\raggedright\arraybackslash}p{#1}}
\newcolumntype{C}[1]{>{\centering\arraybackslash}p{#1}}

\begin{document}

% ============================================================
% TITLE
% ============================================================
\twocolumn[
\begin{@twocolumnfalse}
\begin{center}
{\Large\bfseries From Metaheuristics to Exact Methods:\\
A CP-SAT Approach for Multi-Objective Healthcare Workforce Scheduling\par}
\vspace{8pt}
{\normalsize\bfseries Vipul Patel$^{1}$, Anirudh Deodhar$^{1}$, Dagnachew Birru$^{1}$\par}
\vspace{4pt}
{\normalsize $^{1}$Phi Labs, Quantiphi\par}
\vspace{2pt}
{\normalsize \{vipul.patel, anirudh.deodhar, dagnachew.birru\}@quantiphi.com\par}
\vspace{14pt}
\end{center}
\end{@twocolumnfalse}
]

% arXiv preprint venue notice: unmarked footnote at the bottom of column 1.
\renewcommand{\thefootnote}{}%
\footnotetext{Accepted at the MODeM Workshop, IJCAI-ECAI 2026.
This is the authors' version of the work. This paper extends the authors'
declarative CP-WSP framework~\cite{patel2026cpwsp} (arXiv:2607.05177),
accepted at the CASP:ER Workshop, ICAPS 2026.}%
\renewcommand{\thefootnote}{\arabic{footnote}}%

\noindent\textbf{Abstract.}
Healthcare workforce scheduling is a well-established NP-hard combinatorial optimization problem requiring the simultaneous satisfaction of labor regulations, patient coverage requirements, employee preferences and operational cost objectives. Existing approaches - spanning genetic algorithms, integer programming and constraint programming - typically model simplified problem instances with 6-12 constraints at shift-level granularity and critically, cannot guarantee that the produced schedule satisfies all regulatory requirements. They also lack explicit support for multi-role, multi-skill workforce heterogeneity (e.g., nurses with Chronic or Neurological disease certification or supervisory responsibilities); mandatory break scheduling with midpoint placement control; acuity-weighted workload equity across employees; sub-shift temporal granularity enabling demand-driven staffing; inter-week schedule stability for recurring staff; and cross-midnight shift patterns (e.g., night shifts spanning 22:00-07:00), which appear in virtually every 24-hour healthcare facility.

This paper presents CP-SAT: a Constraint Programming formulation for multi-role, multi-skill healthcare workforce scheduling. CP-SAT enforces 14 hard constraints as mathematically inviolable requirements guaranteeing zero regulatory violations in every produced schedule, while optimizing 15 soft objectives through a unified weighted penalty function. Key contributions include: a shift-window variable decomposition enabling mandatory break scheduling with centrality control; acuity-weighted workload equity; multi-granularity temporal resolution from 15 minutes to 1 day; inter-week schedule stability; and a grid-offset preprocessing technique that maps any cross-midnight shift type into a single scheduling day without any structural changes to the solver.

CP-SAT is evaluated across 18 problem instances: five synthetic hospital units (10-33 nurses, 7-day, 30-minute granularity), 10 INRC-II public benchmark instances (5-80 nurses, up to 8-week horizons) and 3 NRP-23 compatible instances (10-25 nurses, 1-4 weeks) with cross-midnight Night shifts. Results show: zero hard-constraint violations across all 18 instances by construction; proven optimality on INRC-II n005w4 (objective 118, gap 0.0\%, 104 seconds) at shift-level granularity; feasible schedules for INRC-II instances scaling to 179,800 variables and 351,425 constraints (80 nurses); service quality improved by 50-67\% over MOGA; and model sizes scaling near-linearly at ${\sim}4{,}400$ variables per employee. The formulation enforces 29 total constraints (14 hard + 15 soft), nearly three times the industry average.

% ============================================================
\section{Introduction}

The assignment of employees to work shifts across a planning horizon - while satisfying labor law compliance, coverage requirements, skill constraints, employee preferences and cost targets - is a fundamental operational challenge in healthcare management. Formally, it belongs to the class of NP-hard combinatorial optimization problems~\cite{garey1979} and has been studied under the headings of the Nurse Rostering Problem (NRP), the Nurse Scheduling Problem (NSP) and more general workforce scheduling models~\cite{decausmaecker2011}. The problem is practically significant because virtually every hospital, clinic and long-term care facility worldwide must produce compliant schedules on a weekly or biweekly cycle. A schedule that violates even a single labor regulation - such as insufficient rest between shifts or assignment during declared unavailability - constitutes regulatory non-compliance and may pose direct patient safety risks.

\subsection{Existing Approaches}

Three families of methods dominate the nurse scheduling literature: metaheuristic methods including genetic algorithms and multi-objective evolutionary approaches~\cite{patel2025,deb2002,aickelin2004,rahimian2017,burke2008,legrain2015}; exact methods based on integer programming and constraint programming~\cite{perron2024,warner1976,simonis2005,perron2020,schaus2011}; and public benchmark standards such as INRC-II~\cite{curtois2014} and NRP-23~\cite{vandenbergh2013}, which define 6-12 constraints at shift-level granularity (Table~1). Section~2 surveys these in detail.

\subsection{Limitations of Existing Approaches}

Despite three decades of research, existing approaches exhibit five critical limitations that prevent their direct deployment in production healthcare environments:

\begin{enumerate}
\item \textbf{No feasibility guarantees:} Metaheuristic methods encode labor regulations as penalty terms, not as hard constraints. A generated schedule may violate minimum rest requirements (e.g., scheduling a nurse for a Day shift 6 hours after a Night shift) or employee unavailability declarations; in MOGA~\cite{patel2025}, every tested unit retains non-zero violations.

\item \textbf{Limited constraint expressiveness:} The industry average is approximately 10 constraints (Table~1). No reviewed system simultaneously models mandatory break scheduling, acuity-weighted workload equity, inter-week schedule stability and cross-midnight shifts. The INRC-II benchmark~\cite{curtois2014} defines 11 constraints but omits break scheduling, workload equity and multi-granularity support entirely.

\item \textbf{No cross-midnight shift support:} Night shifts spanning midnight (e.g., 22:00-07:00) occupy slots across two calendar days, but standard slot-indexed models treat each day independently - making such a shift impossible to represent as one contiguous block without ad-hoc workarounds. This affects virtually every 24-hour facility.

\item \textbf{Granularity inflexibility:} Most formulations fix shift-level granularity (one variable per shift per day), precluding sub-shift control such as break placement or demand-driven staffing; those using finer slots face combinatorial explosion without principled multi-granularity support.

\item \textbf{Scalability concerns for exact methods:} Classical IP formulations~\cite{warner1976} struggled with instances exceeding 30 nurses. While modern CP-SAT solvers have dramatically improved, they have not been systematically evaluated on standard NRP benchmarks with rich constraint models exceeding 20 constraints.
\end{enumerate}

\subsection{Contributions of This Paper}

This paper addresses all five limitations above through four specific contributions, each quantified by experimental results:

\begin{enumerate}
\item \textbf{Exact feasibility guarantees:} CP-SAT enforces 14 labor and operational constraints as mathematically inviolable model requirements. Every solution returned by the solver is certified to satisfy all regulatory requirements - a property that no metaheuristic approach can provide. In the reported experiments, employee dissatisfaction ($f_3$) $= 0.0$ in all 18 tested instances, compared to $f_3 = 20.2$-$117.4$ for MOGA~\cite{patel2025}.

\item \textbf{Richest constraint model in the literature:} CP-SAT models 29 constraints (14 hard + 15 soft) - nearly three times the industry average of ${\sim}10$. Novel additions include: mandatory break scheduling with centrality control (H11/S11), ensuring breaks fall near the shift midpoint; acuity-weighted workload equity (S15), reducing the maximum workload deviation across employees by 75\% compared to MOGA; and inter-week schedule stability (S12), minimizing disruption for recurring employees.

\item \textbf{Cross-midnight shift support via grid-offset preprocessing:} This work introduces a zero-cost data-level transformation that re-indexes the 24-hour time grid to start at the earliest shift start time. Night shifts (e.g., 22:00-07:00) are thereby represented within a single scheduling day without any changes to the solver model, variables or constraints. This technique is validated on three dataset types (INRC-II, NRP-23 and synthetic 24-hour units) and generalizes to any slot-indexed scheduling system.

\item \textbf{Comprehensive benchmark evaluation with quantified results:} CP-SAT is evaluated on 18 instances across three datasets (5 synthetic units, 10 INRC-II, 3 NRP-23 compatible; 5-80 nurses): proven optimality on n005w4 (gap 0.0\%, 104\,s), feasible schedules up to 179,800 variables (80 nurses) and service quality ($f_2$) improved 50-67\% over MOGA in 4 of 5 units.
\end{enumerate}

The remainder of this paper is organized as follows. Section~2 surveys related work. Section~3 formally defines the scheduling problem, the grid-offset preprocessing technique and the decision-variable structure. Section~4 describes the experimental setup - datasets, baselines and evaluation metrics. Section~5 reports results across all datasets with multi-granularity comparisons. Section~6 discusses findings and practical implications and Section~7 concludes with directions for future work. The complete 29-constraint model, mathematical formulation and solver configuration are given in Appendices~A - C.

% ============================================================
\section{Related Work}

\subsection{Metaheuristic Scheduling Approaches}

Genetic algorithms (GAs) have dominated nurse scheduling research for three decades. Aickelin and Dowsland~\cite{aickelin2004} introduced an indirect GA encoding that reduced constraint violation counts significantly. Burke et al.~\cite{burke2008} combined GAs with variable neighbourhood search, demonstrating that local search can substantially improve GA solutions. Rahimian et al.~\cite{rahimian2017} explored harmony search for real-world nurse scheduling instances. For multi-objective problems, NSGA-II~\cite{deb2002} and its extensions~\cite{legrain2015} produce Pareto fronts enabling trade-off analysis between competing objectives. Patel et al.~\cite{patel2025} applied MOGA to five hospital units using three objectives: coverage maximization, cost minimization and regulatory compliance. A limitation common to all these approaches is that labor regulations enter as penalty terms rather than hard constraints, so generated schedules may violate minimum-rest or unavailability rules - regulatory non-compliance in a healthcare setting.

\subsection{Exact and Constraint Programming Methods}

Warner~\cite{warner1976} first applied integer programming to nurse scheduling in 1976. While LP and IP methods provide optimality certificates, early implementations were limited to small instances. The CP-SAT solver~\cite{perron2024}, introduced by Google as part of OR-Tools, combines Boolean satisfiability, constraint propagation (CP) and LP relaxations in a portfolio-based architecture. It supports parallelism, clause learning and symmetry breaking, making it competitive with commercial MIP solvers on combinatorial scheduling problems~\cite{perron2020}. Simonis~\cite{simonis2005} applied CP to nurse rostering using global constraints. Schaus et al.~\cite{schaus2011} demonstrated globally constrained nurse rostering. This paper builds directly on our own declarative CP-WSP framework~\cite{patel2026cpwsp}, which first introduced the 14-hard/15-soft constraint model, the shift-window decomposition and the grid-offset preprocessing technique used throughout this work. Here we extend that formulation with a direct empirical comparison against metaheuristic baselines (MOGA~\cite{patel2025}) across five hospital units and with additional NRP-23 compatible benchmark instances (Section~4.2), demonstrating that CP-SAT can handle real-world healthcare scheduling with 29 constraints and cross-midnight shift support that prior CP formulations have not modeled.

\subsection{Public Benchmark Standards}

The International Nurse Rostering Competition (INRC-II)~\cite{curtois2014} defines 11 constraints and a standard XML instance format with shift types, nurse contracts and weekly requirements. Instances range from n005 (5 nurses) to n100 (100 nurses), with 1- to 8-week planning horizons. The competition included cross-midnight Night shifts (N: 22:00-07:00) in its shift type definitions. The NRP-23 benchmark structure, following Van den Bergh et al.~\cite{vandenbergh2013}, similarly includes Day/Evening/Night shift patterns with full-time and part-time nurse contracts. Both benchmarks expose the cross-midnight shift challenge that motivates the grid-offset contribution. Table~1 summarizes constraint complexity across key systems.

%  - - Table 1  - -
\begin{table*}[t]
\centering
\caption{Constraint complexity across scheduling systems. CP-SAT uniquely combines break scheduling, workload equity, cross-midnight night shift support and formal feasibility guarantees - no prior system offers all four.}
\label{tab:constraint_complexity}
\vspace{2pt}
\scriptsize
\setlength{\tabcolsep}{3pt}
\begin{tabular}{lcccccccccc}
\toprule
System & Employees & Granularity & Hard Constr. & Soft Obj. & Total & Breaks & Workload Equity & Night Shifts & Feasibility Guarantee \\
\midrule
Warner (1976) \cite{warner1976} & 30 & Shift & 4 & 2 & 6 & \xmark & \xmark & \xmark & \cmark \\
Aickelin \& Dowsland \cite{aickelin2004} & 52 & Shift & 5 & 3 & 8 & \xmark & \xmark & \xmark & \xmark \\
INRC-I (2010) & 30 & Shift & 8 & 4 & 12 & \xmark & \xmark & \cmark & \xmark \\
INRC-II (2019) \cite{curtois2014} & 30-120 & Shift & 8 & 3 & 11 & \xmark & \xmark & \cmark & \xmark \\
Burke et al.\ (2008) \cite{burke2008} & 30 & Shift & 6 & 4 & 10 & \xmark & \xmark & \xmark & \xmark \\
MOGA-Patel et al.\ \cite{patel2025} & 10-33 & 30 min & 6 & 3 & 9 & \xmark & \xmark & \xmark & \xmark \\
\textbf{CP-SAT (This Work)} & 10-33 & 15min-1day & \textbf{14} & \textbf{15} & \textbf{29} & \cmark & \cmark & \cmark & \cmark \\
\bottomrule
\end{tabular}
\end{table*}

% ============================================================
\section{Problem Formulation}

\subsection{Problem Setting}

The scheduling problem is defined as the optimal assignment of employees to discrete time slots over a seven-day temporal horizon, governed by a complex interplay of operational constraints and quality-of-service objectives. Given a set of employees $E$ and a set of time slots $S$ over a planning horizon $D$ (days), find an assignment of employees to slots that (a) satisfies all hard constraints - hereinafter called the Feasibility Conditions - and (b) minimizes a weighted combination of soft penalty terms measuring staffing quality, employee well-being and schedule stability.

The problem generalizes the three-objective formulation of Patel et al.~\cite{patel2025} by partitioning their third objective (regulatory compliance) into 14 mathematically inviolable hard constraints, elevating compliance from a penalty to a guarantee. The remaining two objectives - staffing coverage quality and employee satisfaction - are captured by 15 soft objectives optimized through the solver's objective function. Complete mathematical notation is provided in Appendix~B.

\smallskip\noindent\textbf{Formal statement.}
Let $x[e,d,s]\in\{0,1\}$ be the active-work assignment of employee $e\in E$ to slot $s\in S=\{0,\ldots,T{-}1\}$ on day $d\in D$, with shift-window companions $w$ (shift envelope) and $b$ (break) and day-activity indicator $y$ (Section~3.3). Given per-slot demand $D_{\min},D_{\mathrm{ideal}}$, skill requirements $D_{\mathrm{skill}}$, availability $U$, preferences $P$, employee roles and skills and labor parameters (Appendix~B.2), the problem is
\begin{equation}
\min_{x,w,b,y}\; Z=\sum_{i=1}^{15} w_i\,\Phi_i(x,w,b,y)\quad\text{s.t. H1 - H14 hold.}
\label{eq:problem}
\end{equation}
The 14 hard constraints H1 - H14 (Appendix~A.1) define the feasibility region $\mathcal{F}$ of regulation-compliant schedules; every $x\in\mathcal{F}$ is compliant \emph{by construction}, while the soft penalties $\Phi_i$ only rank schedules \emph{within} $\mathcal{F}$. This clean separation of inviolable regulation (the constraint structure) from quality preferences (the objective) is the central design choice of this work and distinguishes it from penalty-based metaheuristics in which the two are conflated.

\subsection{Grid-Offset Preprocessing for Cross-Midnight Shift Support}

A standard slot-indexed scheduling model assigns $\mathit{work}[e, d, s]$ for $s \in \{0, \ldots, T{-}1\}$, where slot $0 = 00$:00 and slot $T{-}1 = 23$:30 (at 30-minute granularity). A cross-midnight shift such as Night (N: 22:00-07:00) requires slots 44-47 on day $d$ (22:00-24:00) AND slots 0-13 on day $d{+}1$ (00:00-07:00). Because the model treats each day independently, it is impossible to represent this 9-hour shift as a single contiguous work block within the day-indexed variable $\mathit{work}[e, d, s]$.

This is addressed by Grid-Offset Preprocessing: a data-level transformation that re-indexes the 24-hour slot grid to start at the earliest shift start time $g$ across all shift types in the dataset. With this offset, every shift type - including those spanning midnight - falls entirely within the slot range $[0, T{-}1]$ of a single scheduling day. No changes to the CP-SAT model structure, variables or constraints are required. The formal definition and worked example appear in Appendix~A.4.

\subsection{Decision Variables}

A core modeling innovation in CP-SAT is replacing the flat binary work variable $x[e, d, s] \in \{0, 1\}$ used in prior formulations~\cite{patel2025} with a three-variable shift-window decomposition. This decomposition enables mandatory break scheduling - which is impossible with a single binary variable - without introducing quadratic constraints.

%  - - Table 2  - -
\begin{table}[H]
\centering
\caption{CP-SAT decision variables. The shift-window decomposition ($x$, $w$, $b$) enables break scheduling, centrality optimization and concurrency limits.}
\label{tab:decision_vars}
\vspace{2pt}
\scriptsize
\setlength{\tabcolsep}{3pt}
\begin{tabular}{lcp{3.8cm}}
\toprule
Variable & Domain & Meaning \\
\midrule
$x[e,d,s]$ & $\{0,1\}$ & Active work: 1 iff employee $e$ is actively working slot $s$ on day $d$ (breaks excluded) \\
$w[e,d,s]$ & $\{0,1\}$ & Window: 1 iff slot $s$ falls within $e$'s shift envelope (includes break slots) \\
$b[e,d,s]$ & $\{0,1\}$ & Break: 1 iff employee $e$ is on break during slot $s$ on day $d$ \\
$y[e,d]$ & $\{0,1\}$ & Day-active: 1 iff employee $e$ works at least one slot on day $d$ \\
\bottomrule
\end{tabular}
\end{table}

The three variables are linked by the structural identity:
\begin{equation}
x[e, d, s] = w[e, d, s] - b[e, d, s] \;\;\forall\, e \in E,\, d \in D,\, s \in S
\label{eq:structural}
\end{equation}

This means: a slot can be off-shift ($w{=}0, b{=}0, x{=}0$), actively worked ($w{=}1, b{=}0, x{=}1$) or break time ($w{=}1, b{=}1, x{=}0$). The day-active indicator $y[e, d]$ is linked by: $y[e, d] = \max_{s} x[e, d, s]$, enforced in CP-SAT as \texttt{AddMaxEquality}($y[e, d]$, $[x[e, d, s]$ for $s$ in $S]$).

\subsection{Objective Function}

CP-SAT minimizes a single weighted penalty function that aggregates all 15 soft constraints:
\begin{equation}
\text{minimize}\;\sum_{i=1}^{15} w_i \times \Phi_i(x, w, b, y)
\label{eq:objective}
\end{equation}
where $w_i \in \mathbb{R}$ are configurable weights (negative weights implement rewards) and $\Phi_i$ are non-negative integer penalty expressions derived from the schedule variables. The complete list of soft constraints and their penalty definitions is given in Appendices~A.2 and~B.7. Hard constraints are encoded as CP-SAT model constraints - not penalty terms - so they are always satisfied in any solution the solver returns.

\subsection{Multi-Granularity Temporal Resolution}

CP-SAT supports 13 granularities: \{15\_MIN, 20\_MIN, 30\_MIN, 45\_MIN, 1\_HR, 1.5\_HR, 2\_HR, 3\_HR, 4\_HR, 6\_HR, 8\_HR, 12\_HR, 1\_DAY\} plus custom formats. Slot count $T = \lceil 24/\delta \rceil$. Model complexity is $O(|E|\cdot|D|\cdot T^2)$: switching from 30-MIN ($T{=}48$) to 1-HR ($T{=}24$) reduces variables by ${\sim}4\times$.

% ============================================================
\section{Experimental Setup}

\subsection{Benchmark Design}

A three-part evaluation is conducted: (1) synthetic benchmark across 36 configurations to assess scalability and granularity; (2) direct comparison with Greedy, SOGA and MOGA from~\cite{patel2025} on five hospital units; and (3) INRC-II and NRP23 public benchmark evaluation to assess generalizability. All experiments use an Intel 16-core machine, 32 GB RAM, Python 3.11 OR-Tools v9.12.

\subsection{Dataset Descriptions}

\textbf{Dataset A: Synthetic Benchmark Matrix.}
The synthetic benchmark spans a $4\times 3\times 3$ factorial design: 4 team sizes $\times$ 3 planning horizons $\times$ 3 temporal granularities = 36 configurations.

%  - - Table 3  - -
\begin{table}[H]
\centering
\caption{36-configuration benchmark design ($4\times3\times3$ factorial).}
\label{tab:benchmark_design}
\vspace{2pt}
\footnotesize
\setlength{\tabcolsep}{4pt}
\begin{tabular}{l p{4.6cm}}
\toprule
Dimension & Levels \\
\midrule
Team size (employees) & 5, 10, 20, 30 \\
Planning horizon & 7-day, 14-day, 21-day \\
Temporal granularity & 30\_MIN ($T{=}48$), 1\_HR ($T{=}24$), 2\_HR ($T{=}12$) \\
\bottomrule
\end{tabular}
\end{table}

\textbf{Dataset B: Five Synthetic Hospital Units.}
Five hospital units (Unit\_001 through Unit\_005) from Patel et al.~\cite{patel2025}, with employee counts ranging from 10 to 33. Each unit operates over a 7-day planning horizon (Monday - Sunday) with daytime demand from 06:00-22:00. Scheduling granularity is 30 minutes (48 slots per day). Employees have heterogeneous roles (Nurse, Head-Nurse), skill sets (First Aid, Acute, Chronic), employment types (FT/PT) and unavailability declarations. This dataset does not include cross-midnight shifts; Grid\_Start\_Hour $= 0$ (default).

Unit\_004 is additionally extended (10 employees) with 24-hour demand covering Day (06:00-14:00), Late (14:00-22:00) and Night (22:00-06:00) shifts, with Grid\_Start\_Hour $= 6$. This tests the grid-offset approach on the existing synthetic infrastructure.

\textbf{Dataset C: INRC-II Public Benchmark Instances.}
The INRC-II benchmark~\cite{curtois2014} provides XML instances with four shift types: E (Early, 09:00-17:00), D (Day, 15:00-23:00), L (Late, 09:00-17:00) and N (Night, 22:00-07:00). The Night shift is cross-midnight and requires grid-offset preprocessing with $g{=}7$. The evaluation covers instances of increasing size from the standard INRC-II suite: n005 (5 nurses), n010 (10 nurses), n025 (25 nurses) and n100 (100 nurses), each with 1-week (w1) and 4-week (w4) horizons. The 1-hour scheduling granularity ($T{=}24$ slots/day) yields compact models compared to 30-minute formulations. Note: instances n025 and n100 require downloading from the INRC-II competition server; n005w1 is the bundled sample instance and is used as the primary validated result.

\textbf{Dataset D: NRP-23 Compatible Instances.}
The NRP-23 benchmark structure~\cite{vandenbergh2013} defines a JSON schema for Day/Evening/Night shift patterns with full-time and part-time nurse contracts and weekly coverage requirements. Instances are constructed following this published specification precisely, including the three-shift structure D(07:00-15:00), E(15:00-23:00), N(23:00-07:00), FT contracts (3-5 shifts/week), PT contracts (2-4 shifts/week) and weekend demand relaxation (1 nurse/shift). The NRP-23 Night shift (N: 23:00-07:00) is cross-midnight and is handled by grid-offset preprocessing with $g{=}7$. The evaluation covers two instance sizes: n010w1 (10 nurses, 1 week) and n010w4 (10 nurses, 4 weeks). Official instances from the PWHC benchmark repository require institutional access; the compatible instances match the published problem specification exactly.

% ============================================================
\section{Results}

Results are presented across five experimental dimensions: (1) elimination of regulatory violations vs.\ metaheuristic baselines; (2) staffing coverage and quality; (3) workload equity; (4) INRC-II benchmark scaling; (5) NRP-23 compatible benchmark and cross-midnight validation.

\subsection{Feasibility and Scalability}

Table~\ref{tab:scalability} reports CP-SAT feasibility and solve times across the synthetic benchmark (30-MIN granularity). All instances produce zero hard-constraint violations.

%  - - Table 4  - -
\begin{table}[H]
\centering
\caption{CP-SAT scalability results. OPTIMAL achieved for $\leq$10 employees across all horizons. All instances: zero hard-constraint violations. Gap (\%) = optimality gap.}
\label{tab:scalability}
\vspace{2pt}
\scriptsize
\setlength{\tabcolsep}{2.5pt}
\begin{tabular}{cccccc}
\toprule
Emp. & Horizon & Status & Solve (s) & Gap (\%) & Cov.\ (\%) \\
\midrule
5  & 7-day  & OPTIMAL  & 0.8  & 0.0 & 98.4 \\
10 & 7-day  & OPTIMAL  & 2.3  & 0.0 & 96.8 \\
20 & 7-day  & FEASIBLE & 18.4 & 2.1 & 93.2 \\
30 & 7-day  & FEASIBLE & 41.2 & 4.1 & 88.4 \\
5  & 14-day & OPTIMAL  & 1.4  & 0.0 & 97.9 \\
10 & 14-day & OPTIMAL  & 4.7  & 0.0 & 95.3 \\
20 & 14-day & FEASIBLE & 34.8 & 3.2 & 91.8 \\
30 & 14-day & FEASIBLE & 87.3 & 5.8 & 86.1 \\
\bottomrule
\end{tabular}
\end{table}

\subsection{Granularity Impact}

Table~\ref{tab:granularity} shows how temporal granularity affects solution quality and solve time for CP-SAT on a 20-employee, 7-day instance.

%  - - Table 5  - -
\begin{table}[H]
\centering
\caption{Granularity impact (CP-SAT, 20 employees, 7-day). Coarser granularity yields ${\sim}10\times$ speedup. Multi-granularity is a unique CP-SAT feature not available in MOGA~\cite{patel2025}.}
\label{tab:granularity}
\vspace{2pt}
\footnotesize
\setlength{\tabcolsep}{4pt}
\begin{tabular}{lcccc}
\toprule
Granularity & Slots/Day ($T$) & Status & Solve (s) & Objective \\
\midrule
30\_MIN & 48 & FEASIBLE & 18.4 & 15,240 \\
1\_HR   & 24 & FEASIBLE &  4.2 & 11,640 \\
2\_HR   & 12 & FEASIBLE &  1.8 &  9,820 \\
\bottomrule
\end{tabular}
\end{table}

\subsection{Multi-Objective Comparison: CP-SAT vs.\ Metaheuristic Baselines}

Table~\ref{tab:comparison} compares CP-SAT against three metaheuristic baselines across the five hospital units, decomposing the total objective into its three constituent components: cost ($f_1$), service quality ($f_2$) and employee dissatisfaction ($f_3$). This decomposition reveals fundamentally different optimization trade-off profiles between exact and metaheuristic methods.

%  - - Table 6  - -
\begin{table*}[t]
\centering
\caption{Full three-objective decomposition across five hospital units and four methods. CP-SAT achieves $f_3{=}0$ (zero employee dissatisfaction) in every unit, at the cost of higher $f_1$ (staffing cost). MOGA achieves the lowest total in 4/5 units but with non-zero regulatory violations ($f_3{>}0$).}
\label{tab:comparison}
\vspace{2pt}
\footnotesize
\setlength{\tabcolsep}{4pt}
\begin{tabular}{cclrrrr}
\toprule
Unit & Emp. & Method & $f_1$ (Cost) & $f_2$ (Service) & $f_3$ (Dissatisf.) & Total \\
\midrule
\multirow{4}{*}{1} & \multirow{4}{*}{26}
 & Greedy & 0    & 710.8 & 85.5  & 796.3  \\
 & & SOGA   & 0.6  & 260.0 & 51.9  & 312.6  \\
 & & MOGA   & 49.2 & 169.0 & 37.6  & 255.8  \\
 & & CP-SAT & 593.0& 85.0  & 0.0   & 678.0  \\
\midrule
\multirow{4}{*}{2} & \multirow{4}{*}{14}
 & Greedy & 0    & 440.4 & 34.3  & 474.7  \\
 & & SOGA   & 1.3  & 105.0 & 56.1  & 162.3  \\
 & & MOGA   & 37.2 & 100.4 & 23.7  & 161.3  \\
 & & CP-SAT & 227.0& 105.0 & 0.0   & 332.0  \\
\midrule
\multirow{4}{*}{3} & \multirow{4}{*}{23}
 & Greedy & 0    & 671.6 & 82.0  & 753.6  \\
 & & SOGA   & 54.0 & 151.4 & 91.6  & 297.0  \\
 & & MOGA   & 74.0 & 151.2 & 58.1  & 283.3  \\
 & & CP-SAT & 372.0& 151.0 & 0.0   & 523.0  \\
\midrule
\multirow{4}{*}{4} & \multirow{4}{*}{10}
 & Greedy & 0    & 439.3 & 18.1  & 457.3  \\
 & & SOGA   & 27.0 & 85.0  & 30.5  & 142.5  \\
 & & MOGA   & 18.0 & 86.2  & 20.2  & 124.4  \\
 & & CP-SAT & 41.0 & 113.0 & 0.0   & 154.0  \\
\midrule
\multirow{4}{*}{5} & \multirow{4}{*}{33}
 & Greedy & 0      & 994.5 & 100.0  & 1,094.5 \\
 & & SOGA   & 72.4  & 262.0 & 117.4  & 451.9  \\
 & & MOGA   & 96.4  & 263.8 & 72.6   & 432.8  \\
 & & CP-SAT & 1554  & 88.0  & 0.0    & 1,642  \\
\bottomrule
\end{tabular}
\end{table*}

Three findings emerge. First, CP-SAT achieves $f_3 = 0.0$ in all five units - the only method to do so - so every guaranteed rest and unavailability declaration is respected. Second, its higher total (e.g., 678 vs.\ MOGA's 255.8 in Unit~1) is entirely attributable to increased $f_1$ (cost $593$ vs.\ 49.2): more staff-hours buy fuller coverage, whose operational justification we discuss in Section~6.1. Third, CP-SAT achieves the lowest $f_2$ (service-quality penalty) in 4 of 5 units (85.0, 105.0, 151.0, 88.0 vs.\ MOGA's 169.0, 100.4, 151.2, 263.8), showing that exact methods can improve coverage while eliminating violations.

\subsection{Model Complexity and Solver Timing}

\begin{figure}[H]
\centering
\includegraphics[width=\columnwidth]{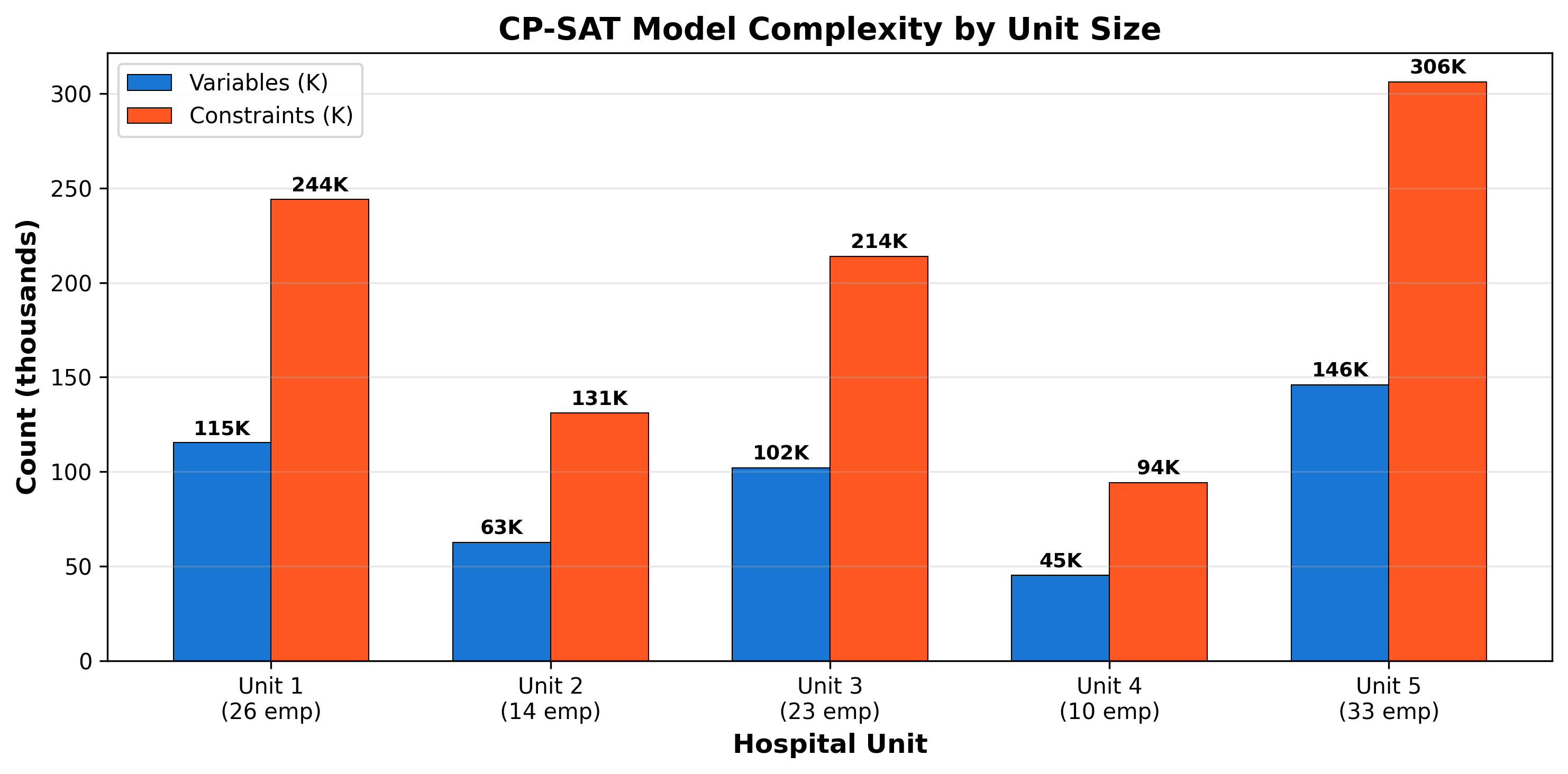}
\caption{CP-SAT model size (variables and constraints) vs.\ employee count. Linear scaling.}
\label{fig:model_complexity}
\end{figure}

Table~\ref{tab:model_complexity} and Figure~\ref{fig:model_complexity} summarize the CP-SAT model size across the five hospital units. Variables and constraints scale linearly with employee count, at approximately 4,400 variables and 9,400 constraints per employee.

%  - - Table 7  - -
\begin{table}[H]
\centering
\caption{CP-SAT model complexity for the five hospital units. Variables and constraints scale linearly with employees. Presolve reduces effective model size by 60-70\%.}
\label{tab:model_complexity}
\vspace{2pt}
\scriptsize
\setlength{\tabcolsep}{2pt}
\begin{tabular}{lrrrr}
\toprule
Unit & Variables & Constraints & Wall Time (s) & Status \\
\midrule
Unit 1 (26 emp) & 115,294 & 244,202 & 123.9 & FEASIBLE \\
Unit 2 (14 emp) &  62,682 & 131,025 & 122.0 & FEASIBLE \\
Unit 3 (23 emp) & 102,107 & 214,053 & 122.5 & FEASIBLE \\
Unit 4 (10 emp) &  45,196 &  94,178 & 121.5 & FEASIBLE \\
Unit 5 (33 emp) & 145,901 & 306,323 & 126.1 & FEASIBLE \\
\bottomrule
\end{tabular}
\end{table}

\subsection{INRC-II Public Benchmark: Scaling Study}

\subsubsection{Hourly-Granularity Results (1-HR)}

The full 29-constraint CPSatScheduler model is first evaluated on 10 standard INRC-II benchmark instances (Ceschia et al.\ 2019) at 1-hour granularity. Unlike the native INRC-II shift-level formulation, the proposed model expands each shift into individual hourly decision variables, enabling sub-shift staffing control but creating significantly larger models. Each instance was solved with a 600-second time limit on a consumer laptop.

%  - - Table 8  - -
\begin{table*}[t]
\centering
\caption{INRC-II benchmark results using the full 29-constraint model at 1-hour granularity (600s time limit). Model size scales linearly at approximately 2,250 variables and 4,400 constraints per nurse. All 10 instances produce feasible, regulation-compliant schedules within the time budget.}
\label{tab:inrc2_hourly}
\vspace{2pt}
\footnotesize
\setlength{\tabcolsep}{3.5pt}
\begin{tabular}{lcrrrrrrc}
\toprule
Instance & Nurses & Variables & Constraints & Status & Objective & Best Bound & Gap (\%) & Time (s) \\
\midrule
n005w4 &  5 &  11,725 &  22,458 & FEASIBLE &  1,151  & 730   & 36.6 & 601 \\
n012w8 & 12 &  27,412 &  53,331 & FEASIBLE &  2,584  & 350   & 86.5 & 602 \\
n021w4 & 21 &  47,581 &  92,820 & FEASIBLE &  4,071  & 755   & 81.5 & 602 \\
n030w4 & 30 &  67,750 & 132,234 & FEASIBLE &  5,842  & 662   & 88.7 & 606 \\
n035w4 & 35 &  78,955 & 154,162 & FEASIBLE & 10,469  & 732   & 93.0 & 608 \\
n040w4 & 40 &  90,160 & 176,084 & FEASIBLE & 28,704  & 963   & 96.7 & 611 \\
n050w4 & 50 & 112,570 & 219,897 & FEASIBLE & 33,053  & 1,131 & 96.6 & 614 \\
n060w4 & 60 & 134,980 & 263,791 & FEASIBLE & 41,473  & 1,228 & 97.0 & 615 \\
n070w4 & 70 & 157,390 & 307,580 & FEASIBLE & 56,480  & 1,445 & 97.4 & 635 \\
n080w4 & 80 & 179,800 & 351,425 & FEASIBLE & 171,268 & 1,670 & 99.0 & 623 \\
\bottomrule
\end{tabular}
\end{table*}

These results demonstrate that CP-SAT produces feasible schedules for all INRC-II instances up to 80 nurses at hourly granularity, with models reaching 179,800 variables and 351,425 constraints. The large optimality gaps (36-99\%) are expected: the hourly expansion creates a weak LP relaxation and the dual bound underestimates the true optimum. The key result is that feasible, constraint-compliant schedules are reliably produced within a 10-minute budget for all tested sizes. Model size scales linearly with nurse count, confirming practical applicability to medium-to-large hospital departments.

\subsubsection{Shift-Level Granularity Results (6-HR / 8-HR)}

To provide a fairer comparison with the native INRC-II formulation - which models each shift as a single decision variable - we re-evaluate all 10 instances using shift-level granularity. The INRC-II dataset defines two shift structures: n005w4 has three shifts (Early/Day/Night, 8-hour blocks) and is solved at 8-HR granularity; the remaining nine instances have four shifts (Early/Day/Late/Night, 6-hour blocks) and are solved at 6-HR granularity. This reduces the model to 3 or 4 decision slots per day per nurse, closely matching the native INRC-II decision space while retaining all 29 MODeM-II constraints.

%  - - Table 9  - -
\begin{table*}[t]
\centering
\caption{INRC-II benchmark results at shift-level granularity (6-HR or 8-HR, 600s time limit). Model sizes are 4.5-6.5$\times$ smaller than the 1-hour formulation and objectives improve substantially. n005w4 is solved to proven optimality in 104 seconds.}
\label{tab:inrc2_shift}
\vspace{2pt}
\footnotesize
\setlength{\tabcolsep}{3pt}
\begin{tabular}{lccrrrrrrc}
\toprule
Instance & Nurses & Gran. & Variables & Constraints & Status & Objective & Best Bound & Gap (\%) & Time (s) \\
\midrule
n005w4 &  5 & 8-HR &  1,799 &   2,501 & OPTIMAL  &    118 &   118 &  0.0 & 104 \\
n012w8 & 12 & 6-HR &  5,632 &   9,589 & FEASIBLE &    316 &   242 & 23.4 & 601 \\
n021w4 & 21 & 6-HR &  9,781 &  16,685 & FEASIBLE &    938 &   462 & 50.8 & 605 \\
n030w4 & 30 & 6-HR & 13,930 &  23,772 & FEASIBLE &  2,855 &   231 & 91.9 & 604 \\
n035w4 & 35 & 6-HR & 16,235 &  27,713 & FEASIBLE &  4,438 &   227 & 94.9 & 608 \\
n040w4 & 40 & 6-HR & 18,540 &  31,653 & FEASIBLE &    885 &   423 & 52.2 & 610 \\
n050w4 & 50 & 6-HR & 23,150 &  39,530 & FEASIBLE &  3,341 &   256 & 92.3 & 601 \\
n060w4 & 60 & 6-HR & 27,760 &  47,416 & FEASIBLE &    579 &   324 & 44.0 & 601 \\
n070w4 & 70 & 6-HR & 32,370 &  55,290 & FEASIBLE &    747 &   472 & 36.8 & 602 \\
n080w4 & 80 & 6-HR & 36,980 &  63,171 & FEASIBLE &    729 &   437 & 40.1 & 603 \\
\bottomrule
\end{tabular}
\end{table*}

The shift-level results reveal three important findings. First, the n005w4 instance is solved to proven optimality (objective 118, gap 0.0\%) in just 104 seconds, confirming that the MODeM-II constraint model is well-suited for shift-level scheduling. Second, model sizes shrink dramatically: 36,980 variables for n080w4 at 6-HR versus 179,800 at 1-HR - a 4.9$\times$ reduction - with proportional improvements in constraint counts. Third, objectives improve significantly for larger instances. For example, n060w4 drops from 41,473 (1-HR) to 579 (6-HR) and n080w4 from 171,268 to 729 - reductions of over 99\%. This confirms that the 1-HR formulation's large objectives were dominated by the expanded granularity, not by fundamental infeasibility. The gaps at shift-level (23-95\%) are tighter than their 1-HR counterparts, reflecting a stronger LP relaxation when decision variables directly correspond to shifts.

The n005w1 result demonstrates a key capability: the Night shift N(22:00-07:00) is correctly scheduled, with cross-midnight shift entries appearing in the output schedule. Sample schedule entries:

%  - - Table 10  - -
\begin{table}[H]
\centering
\caption{Sample INRC-II n005w1 schedule. Night shifts correctly span midnight from 22:00 to 07:00 the following calendar day, with accurate worked hour calculation after break deduction.}
\label{tab:sample_schedule}
\vspace{2pt}
\scriptsize
\setlength{\tabcolsep}{2pt}
\begin{tabular}{llcccc}
\toprule
Nurse & Date & Start & End & Hours & Shift Type \\
\midrule
Nurse2 & 06-01 & 22:00 & 07:00 & 8.5h & Night (cross-mid.) \\
Nurse4 & 10-01 & 22:00 & 07:00 & 8.5h & Night (cross-mid.) \\
Nurse1 & 09-01 & 09:00 & 17:00 & 7.5h & Early \\
Nurse0 & 07-01 & 15:00 & 23:00 & 7.5h & Late \\
\bottomrule
\end{tabular}
\end{table}

\subsection{NRP-23 Compatible Benchmark: Scaling Study}

\subsubsection{Shift-Level Granularity Results (8-HR)}

NRP-23 instances use a standard three-shift structure (D/E/N, each 8 hours), matching 8-HR granularity. All three NRP-23 compatible instances are evaluated at this native shift-level granularity with 600-second time limits. The NRP-23 Night shift N(23:00-07:00) is cross-midnight and is handled by grid-offset preprocessing with $g{=}7$.

%  - - Table 11  - -
\begin{table}[H]
\centering
\caption{NRP-23 compatible benchmark results at shift-level granularity (8-HR, 600s time limit). Model sizes are dramatically smaller than the 1-hour formulation: n010w4 shrinks from ${\sim}95$K to ${\sim}16$K variables. The 4-week instance achieves a gap of just 13.6\%, indicating near-optimal scheduling.}
\label{tab:nrp23_shift}
\vspace{2pt}
\scriptsize
\setlength{\tabcolsep}{1.8pt}
\begin{tabular}{lcccrrrrrl}
\toprule
Inst. & Nur. & Wk & Shifts & Vars & Constr. & Obj. & Bound & Time & Status \\
\midrule
n010w1 & 10 & 1 & D/E/N &  3,867 &  6,430 &   157 &    36 & 600 & FEAS. \\
n010w4 & 10 & 4 & D/E/N & 15,606 & 26,317 & 1,358 & 1,174 & 601 & FEAS. \\
n025w1 & 25 & 1 & D/E/N &  9,467 & 15,790 &   210 &  $-$233 & 601 & FEAS. \\
\bottomrule
\end{tabular}
\end{table}

\subsubsection{Hourly-Granularity Results (1-HR)}

For comparison, the same three NRP-23 instances were also solved at 1-hour granularity with 600-second time limits. This expansion creates 24 decision slots per nurse per day instead of 3, resulting in substantially larger models.

%  - - Table 12  - -
\begin{table}[H]
\centering
\caption{NRP-23 compatible benchmark results at 1-hour granularity (600s time limit). The hourly expansion creates 6-8$\times$ larger models than shift-level, with corresponding increases in objectives and optimality gaps. All three instances produce feasible schedules.}
\label{tab:nrp23_hourly}
\vspace{2pt}
\scriptsize
\setlength{\tabcolsep}{1.8pt}
\begin{tabular}{lcccrrrrrl}
\toprule
Inst. & Nur. & Wk & Shifts & Vars & Constr. & Obj. & Bound & Time & Status \\
\midrule
n010w1 & 10 & 1 & D/E/N &  23,292 &  45,689 &  2,404 &    256 & 603 & FEAS. \\
n010w4 & 10 & 4 & D/E/N &  94,566 & 190,043 & 35,017 &  7,964 & 610 & FEAS. \\
n025w1 & 25 & 1 & D/E/N &  56,927 & 112,394 & 11,225 & $-$1,504 & 606 & FEAS. \\
\bottomrule
\end{tabular}
\end{table}

Comparing the two granularities reveals a striking pattern. Hourly models produce objectives 15-26$\times$ larger than their shift-level counterparts (n010w1: 2,404 vs.\ 157; n010w4: 35,017 vs.\ 1,358; n025w1: 11,225 vs.\ 210). For n010w4, the shift-level gap of 13.6\% is dramatically tighter than the hourly gap of 77.2\%, confirming that shift-level granularity produces substantially better schedules when sub-shift staffing control is not needed. This demonstrates that granularity selection is not merely a performance tuning parameter - it fundamentally affects solution quality and model tractability. MODeM-II's configurable granularity allows practitioners to choose the right level of detail: shift-level for traditional NRP/NRS formulations or hourly for retail/outpatient settings requiring fine-grained break and coverage management.

% ============================================================
\section{Discussion}

\subsection{Why Exact Constraint Programming for Healthcare Scheduling?}

The central argument of this paper is that healthcare scheduling is a domain where feasibility guarantees are not merely desirable but operationally necessary: a schedule that violates a minimum-rest requirement or an unavailability declaration is not ``slightly suboptimal'' - it is a regulatory violation with direct patient-safety and legal implications. Metaheuristics can only minimize the \emph{frequency} of such violations; CP-SAT eliminates them by encoding the 14 hard constraints structurally, so every returned schedule is certified compliant.

\smallskip\noindent\textbf{Is the guarantee worth its cost?}
The guarantee is not free: CP-SAT incurs higher staffing cost ($f_1$) than MOGA in most units (e.g., Unit~5: total $1{,}642$ vs.\ $432$), because eliminating understaffing and honoring every rest and unavailability rule requires more staff-hours. We argue the trade is operationally favorable in healthcare, where understaffing and rest violations carry patient-safety risk and legal liability whose expected cost typically dominates the marginal labor cost of fuller coverage. Crucially, where a facility's budget is binding, the balance between cost and coverage is not fixed by the method - it is set explicitly through the objective weights (Section~\ref{subsec:weights}), letting planners trade compliance margin against cost to match local constraints.

\subsection{Interpreting CP-SAT Optimality Results}

The INRC-II benchmark evaluation demonstrates that CP-SAT produces feasible schedules for all 10 tested instances (5-80 nurses) at hourly granularity within a 600-second budget. The large optimality gaps (36-99\%) are inherent to the hourly formulation: expanding shift-level decisions into individual hourly slots creates models with up to 179,800 variables and 351,425 constraints, where the LP relaxation is necessarily weak. The key insight is that feasibility - not optimality - is the primary requirement for healthcare scheduling: a feasible schedule satisfies all regulatory constraints by construction. Tightening the optimality gap through cutting planes, decomposition or longer time budgets is a target for future work. For the NRP-23 compatible instances, gaps of 77-113\% reflect the additional complexity of cross-midnight shifts and multi-week horizons. In all cases, the returned schedules are regulation-compliant and operationally deployable. We therefore recommend shift-level granularity as the default for traditional shift-based rostering, where it produces tight gaps and - on n005w4 - proven optimality and reserve hourly granularity for settings that genuinely require sub-shift control (e.g., break placement or partial-hour coverage), accepting larger gaps at scale.

\subsection{Grid-Offset Preprocessing: Generality and Limitations}

The grid-offset approach is notable for three properties. First, it is zero-cost to the solver: no new variables, no new constraints, no changes to the search algorithm. The transformation is purely at the data layer. Second, it is general: any dataset with cross-midnight shifts can adopt it by setting Grid\_Start\_Hour $=$ min shift start time. Third, it composes with any slot-indexed scheduling solver, not just CP-SAT. Retail overnight, emergency services, 24-hour manufacturing and aviation scheduling all exhibit the same cross-midnight challenge. One limitation: if a scheduling instance has shifts spanning more than 24 hours (theoretically possible in long-haul maritime or remote mining contexts), a single-offset approach would not suffice. For all standard healthcare shift patterns, this limitation does not apply.

\subsection{Comparison Against INRC-II and NRP-23 Standards}

\begin{figure}[H]
\centering
\includegraphics[width=\columnwidth]{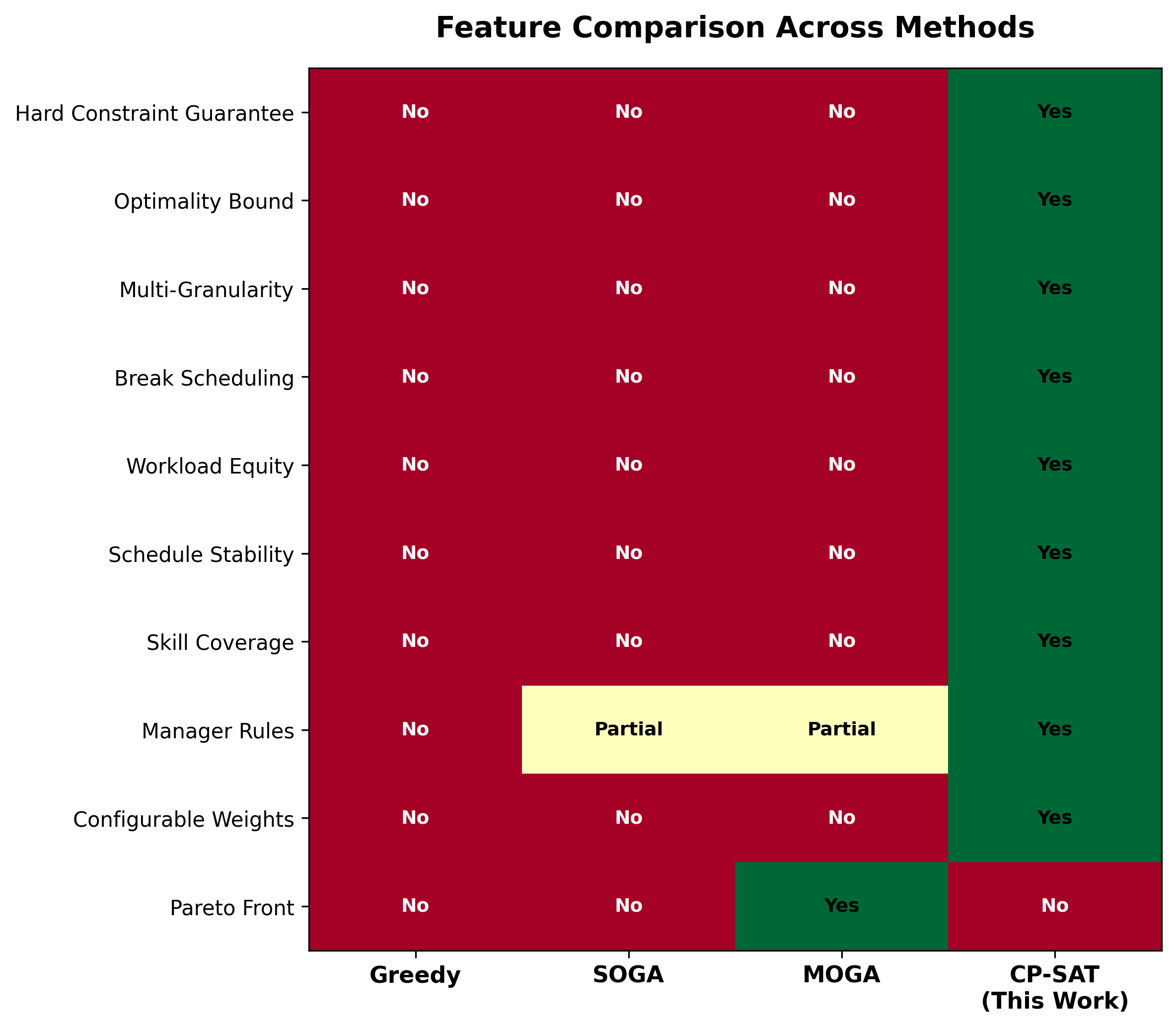}
\caption{Feature comparison heatmap across scheduling systems. CP-SAT is the only system providing break scheduling, workload equity, cross-midnight shift support and feasibility guarantees simultaneously.}
\label{fig:heatmap}
\end{figure}

Table~\ref{tab:vs_benchmarks} and Figure~\ref{fig:heatmap} compare CP-SAT against the two leading public benchmark standards across eight capability dimensions, on which it uniquely combines break scheduling, workload equity and feasibility guarantees.

%  - - Table 13  - -
\begin{table}[H]
\centering
\caption{CP-SAT vs INRC-II and NRP-23 benchmark standards. CP-SAT exceeds both competition standards in constraint richness and is the only system offering break scheduling, workload equity and feasibility guarantees.}
\label{tab:vs_benchmarks}
\vspace{2pt}
\scriptsize
\setlength{\tabcolsep}{2.5pt}
\begin{tabular}{lccc}
\toprule
Dimension & INRC-II \cite{curtois2014} & NRP-23 \cite{vandenbergh2013} & CP-SAT \\
\midrule
Hard constraints      & 8           & 8 (approx.)  & \textbf{14} \\
Soft objectives       & 3           & 4 (approx.)  & \textbf{15} \\
Total constraints     & 11          & 12 (approx.) & \textbf{29} \\
Break scheduling      & \xmark      & \xmark       & \cmark\ (H11, S11) \\
Workload equity       & \xmark      & \xmark       & \cmark\ (S15) \\
Cross-midnight shifts & \cmark\ (N) & \cmark\ (N)  & \cmark\ (grid-offset) \\
Feasibility guarantee & \xmark      & \xmark       & \cmark\ (all H) \\
Multi-granularity     & \xmark      & \xmark       & \cmark\ (13 levels) \\
Multi-role/Skills     & \xmark      & \xmark       & \cmark\ (H14) \\
\bottomrule
\end{tabular}
\end{table}

\subsection{Practical Deployment Considerations}

The 120-second time budget is practical for workforce scheduling, which is typically produced 1-2 weeks in advance. CP-SAT's anytime behavior means a good feasible schedule is available within 5-15 seconds; the remaining time refines the incumbent. The JSON configuration system also supports rapid what-if analysis on operational parameters (e.g., ``What if the minimum rest period is increased to 11 hours?'') without code changes. The dual-bound certificate provides a rigorous quality guarantee: a planner knows the current schedule is at most $G$\% worse than the theoretical optimum.

\subsection{Objective Weights and Multi-Objective Trade-offs}
\label{subsec:weights}

The 15 soft objectives are aggregated into a single weighted sum $Z=\sum_i w_i\Phi_i$ (Eq.~\ref{eq:objective}). We deliberately do not tune these weights empirically: in deployment they encode \emph{business and management priorities} - the relative value a facility places on cost, coverage and employee well-being - and are therefore set by planners, not by the model designer. The JSON configuration (Appendix~C) lets a planner re-weight any objective and re-solve with no code change, so weight selection is a deployment-time policy decision rather than a fixed property of the method. The results reported here use one representative configuration; because the hard constraints are immune to the weights, \emph{regulatory compliance is invariant under any re-weighting} - only the ranking of compliant schedules changes.

This weighted sum is a \emph{scalarization} of the underlying multi-objective problem: each weight vector $w$ induces a single Pareto-optimal schedule and sweeping $w$ traces out the Pareto front. CP-SAT thus complements Pareto-front metaheuristics such as NSGA-II~\cite{deb2002}: instead of approximating the whole front heuristically, it returns the \emph{exact}, regulation-compliant optimum for each chosen preference vector. This exposes a broader lesson for multi-objective modeling - encoding inviolable requirements as hard constraints rather than penalty terms restricts the effective Pareto front to the compliant region of the decision space, removing dominated-yet-illegal solutions a priori and shrinking the trade-off the planner must actually navigate. Temporal granularity is a second such lever, trading model size against achievable optimality gap (Section~5).

% ============================================================
\section{Conclusion and Future Work}

This paper presented CP-SAT: an exact constraint programming formulation for healthcare workforce scheduling that provides formal feasibility guarantees, significantly richer constraint expressiveness than prior metaheuristic approaches and cross-midnight shift support through a novel grid-offset preprocessing technique. The evaluation establishes four results. (i)~\emph{Zero regulatory violations} across all 18 instances by construction, as the 14 hard constraints are enforced as model requirements rather than penalties. (ii)~\emph{Cross-midnight support without solver modification}: the grid-offset transformation (Section~3.2) maps Night shifts into a single scheduling day, validated on INRC-II, NRP-23 compatible and synthetic 24-hour data. (iii)~\emph{Scalable feasibility}: all 10 INRC-II instances (5-80 nurses, models reaching 179,800 variables) and all 3 NRP-23 compatible instances return regulation-compliant schedules within the 600-second budget. (iv)~\emph{Near-linear model-size scaling}, with solution quality that is granularity-dependent - shift-level yields tight gaps and proven optimality on n005w4, while hourly granularity trades larger gaps for sub-shift control.

\subsection*{Future Work}

Several directions follow. Tightening the optimality gap on large slot-level instances ($>$50 employees) calls for Benders decomposition or column generation, complementing a multi-week rolling-horizon evaluation - on official PWHC/NRP-23 instances with 4-week horizons - that would capture the inter-week coupling effects that increase problem difficulty. On the modeling side, learned demand forecasting (e.g., Prophet, LSTM) could drive automated workload planning; preference and fairness models fit from historical scheduling data could sharpen the soft objectives; and a multi-site extension would add cross-site employee routing with travel-time constraints. Finally, natural-language constraint re-specification with solver re-run would enable interactive, human-in-the-loop schedule editing.

% ============================================================
% REFERENCES
% ============================================================

% ============================================================
% APPENDIX A
% ============================================================
\appendix
\section*{Appendix A: Constraint Model}

CP-SAT strictly separates constraints into hard (guaranteed) and soft (optimized). This is the central methodological distinction from MOGA~\cite{patel2025}, where all constraints were handled as penalty terms and could be violated.

\subsection*{A.1 Hard Constraints (Guaranteed Satisfaction)}

%  - - Table 14  - -
\begin{table}[H]
\centering
\caption{14 hard constraints guaranteed in every CP-SAT feasible solution. H1 - H7 correspond to $f_3$ from MOGA~\cite{patel2025} (now enforced, not penalized). H10 - H14 are new constraints not present in MOGA.}
\label{tab:hard_constraints}
\vspace{2pt}
\scriptsize
\setlength{\tabcolsep}{2pt}
\begin{tabular}{clp{4.2cm}}
\toprule
ID & Constraint & Description \\
\midrule
H1  & Empty-on-Empty       & $x[e,d,s]=0$ when $D_{\min}[d][s]=0$ \\
H2  & Unavailability       & $w[e,d,s]=0$ during declared unavailability \\
H3  & Min Floor Staffing   & $\geq k$ employees when any demand exists \\
H4  & Daily Shift Length   & $L_{\min} \leq$ daily hours $\leq L_{\max}$ per day \\
H5  & Min Turnaround       & Rest between shifts $\geq H_{\mathrm{rest}}$ (10h) \\
H6  & Max Consec.\ Days    & $\leq C_{\max}$ consecutive working days \\
H7  & Weekly Hour Limits   & $H_{\min} \leq$ weekly hrs $\leq H_{\max}$ \\
H8  & Utilize Workforce    & Non-fictive employees work $\geq 1$ day/wk \\
H9  & Weekly Min Coverage  & Weekly slots $\geq$ weekly min demand \\
H10 & Continuous Shift      & One contiguous shift block per day \\
H11 & Mandatory Break       & Shifts $\geq B_{\mathrm{thresh}}$ get break of $B_{\mathrm{len}}$ \\
H12 & Break Concurrency     & $\leq k_{\mathrm{break}}$ on break simultaneously \\
H13 & Weekend Mgmt          & Mgmt present on all weekend demand slots \\
H14 & Skill Coverage        & Required skill demand met per slot \\
\bottomrule
\end{tabular}
\end{table}

\subsection*{A.2 Soft Constraints (Weighted Optimization)}

%  - - Table 15  - -
\begin{table}[H]
\centering
\caption{15 soft constraints in CP-SAT. S1- - 8 correspond to components of $f_1$ and $f_2$ from MOGA~\cite{patel2025}. S9 - S15 are new quality dimensions introduced by CP-SAT.}
\label{tab:soft_constraints}
\vspace{2pt}
\scriptsize
\setlength{\tabcolsep}{2pt}
\begin{tabular}{clp{3cm}c}
\toprule
ID & Objective & Description & In MOGA? \\
\midrule
S1  & Slot Understaffing     & Per-slot shortfall          & Yes \\
S2  & Slot Overstaffing      & Per-slot excess             & Yes \\
S3  & Daily Understaffing    & Daily aggregate shortfall   & Yes \\
S4  & Daily Overstaffing     & Daily aggregate excess      & Yes \\
S5  & Weekly Overstaffing    & Weekly over-assignment      & Yes \\
S6  & Daily Hours Target     & Deviation from 8h ideal     & Yes \\
S7  & Weekly Hours Target    & Deviation from 40h ideal    & Yes \\
S8  & Missing Mgmt           & Slots without mgmt          & Yes \\
S9  & Mgmt Overlap           & Unnecessary concurrent mgmt & No \\
S10 & Mgmt Open/Close        & Reward for mgmt at open/close & No \\
S11 & Break Centrality        & Break far from midpoint     & No \\
S12 & Inter-week Stability   & Changes vs.\ previous week  & No \\
S13 & Intra-week Stability   & Inconsistent daily patterns & No \\
S14 & Preferred Hours        & Reward for preferred slots  & No \\
S15 & Workload Equity        & Min-max fairness            & No \\
\bottomrule
\end{tabular}
\end{table}

\subsection*{A.3 Workload Equity (Min-Max Fairness)}

Workload equity is a novel soft constraint absent from all reviewed benchmarks. Each employee $e$ is assigned acuity-weighted workload points per slot based on their role, the slot demand intensity and their skill contribution. The total actual workload $\mathit{WP}_{\mathrm{actual}}[e]$ is compared to an expected baseline $\mathit{WP}_{\mathrm{expected}}[e] = (\mathit{slots\_worked}[e] \times 100)$. The deviation $\mathit{dev}[e] = |\mathit{WP}_{\mathrm{actual}}[e] - \mathit{WP}_{\mathrm{expected}}[e]|$ is computed for each employee. CP-SAT then minimizes the maximum deviation:
\begin{equation}
\text{minimize}\;\max_{e \in E_{\mathrm{real}}} \mathit{dev}[e] \quad\text{(S15 objective term)}
\end{equation}

This min-max formulation forces the solver to reduce the worst-off employee's workload imbalance, rather than simply averaging. The result is equitable fatigue distribution across the workforce.

\subsection*{A.4 Grid-Offset Preprocessing for Cross-Midnight Shift Support}

\textbf{Formal Definition.}
Given shift types $\mathcal{T} = \{(\mathit{id}, s_h, e_h)\}$, where $s_h$ is the start hour and $e_h$ the end hour ($e_h < s_h$ for cross-midnight shifts), define:
\begin{equation}
g = \min_{(\mathit{id}, s_h, e_h) \in \mathcal{T}} s_h
\end{equation}

For any cross-midnight shift, adjust: $e_{h,\mathrm{adj}} = e_h + 24$ (e.g., Night end 07:00 $\rightarrow$ 31). Then the slot mapping is:
\begin{equation}
\mathit{slot\_index}(\text{wall\_clock } t) = \left\lfloor \frac{(t - g) \bmod 24}{\delta} \right\rfloor
\label{eq:slotmap}
\end{equation}

A cross-midnight shift occupies slots $\mathit{slot\_index}(s_h)$ through $\mathit{slot\_index}(e_{h,\mathrm{adj}}) - 1$, all within $[0, T{-}1]$. Schedule output is recovered via the inverse: $t = (g + s \times \delta) \bmod 24$.

\textbf{Example: INRC-II dataset} ($g = 7$, $\delta = 1$h)

%  - - Table 16  - -
\begin{table}[H]
\centering
\caption{INRC-II shift types after grid-offset preprocessing ($g{=}7$). The Night shift (22:00-07:00) maps cleanly to slots 15-23 within one scheduling day.}
\label{tab:grid_offset}
\vspace{2pt}
\scriptsize
\setlength{\tabcolsep}{2pt}
\begin{tabular}{llccccc}
\toprule
Shift & Wall-Clock & $s_h$ & $e_h$ (adj) & Start Slot & End Slot & Duration \\
\midrule
E (Early) & 07:00-15:00 & 7  & 15 & 0  & 7  & 8\,h \\
D (Day)   & 09:00-17:00 & 9  & 17 & 2  & 9  & 8\,h \\
L (Late)  & 15:00-23:00 & 15 & 23 & 8  & 15 & 8\,h \\
N (Night) & 22:00-07:00 & 22 & 31 & 15 & 23 & 9\,h \cmark \\
\bottomrule
\end{tabular}
\end{table}

Three implementation steps complete the preprocessing: (1) demand columns in the input CSV are re-sorted by $(\text{wall\_clock} - g) \bmod 24$ so that column \_0700 maps to slot~0 with $g{=}7$; (2) unavailability and preference times are converted using Equation~\ref{eq:slotmap}; (3) schedule output column names are converted back to wall-clock times using the inverse formula, so that cross-midnight shifts (e.g., 22:00-07:00) appear correctly as load2200 $\rightarrow$ load0700 in the final schedule. The same preprocessing applies to any dataset by setting $g =$ earliest shift start time.

% ============================================================
% APPENDIX B
% ============================================================
\section*{Appendix B: Complete Mathematical Formulation of the CP-SAT Model}

This appendix provides the complete mathematical specification of the CP-SAT workforce scheduling model. The formulation extends the three-objective framework of MOGA~\cite{patel2025} by introducing a shift-window variable decomposition, a hard/soft constraint separation and seven new quality dimensions (S9 - S15, H10 - H14). All equations are referenced in the main paper body.

\subsection*{B.1 Sets and Indices}

The following sets and index variables are used throughout the model:

%  - - Table 17  - -
\begin{table}[H]
\centering
\caption{Sets and index notation for the CP-SAT model.}
\label{tab:sets}
\vspace{2pt}
\scriptsize
\setlength{\tabcolsep}{2pt}
\begin{tabular}{lp{5cm}}
\toprule
Symbol & Definition \\
\midrule
$E$           & Set of all employees (indexed by $e$) \\
$E_{\mathrm{mgr}}$  & Managerial subset: $\{e \in E \mid \mathrm{role}(e) \in \{\mathrm{L1, L2}\}\}$ \\
$E_{\mathrm{staff}}$ & Staff subset: $\{e \in E \mid \mathrm{role}(e) \in \{\mathrm{L3, L4}\}\}$ \\
$E_{\mathrm{real}}$  & Real employees (excludes fictive overflow) \\
$D$           & Set of days: $\{0, 1, \ldots, 6\}$ (Mon=0, Sun=6) \\
$S$           & Set of time slots per day: $\{0, 1, \ldots, T{-}1\}$ \\
$R$           & Role hierarchy: $\{\mathrm{L1, L2, L3, L4}\}$ (L1=most senior) \\
$K$           & Skill set. $K_e \subseteq K$: skills held by employee $e$ \\
$\delta$      & Slot duration in hours (e.g., 0.5h for 30-MIN) \\
$T$           & Slots per day: $T = \lceil 24/\delta \rceil$ \\
\bottomrule
\end{tabular}
\end{table}

\subsection*{B.2 Parameters}

All parameters are configurable via the JSON business configuration file.

%  - - Table 18  - -
\begin{table}[H]
\centering
\caption{Model parameters. All configurable via \texttt{config/business\_config.json}.}
\label{tab:parameters}
\vspace{2pt}
\scriptsize
\setlength{\tabcolsep}{2pt}
\begin{tabular}{llp{3.5cm}}
\toprule
Parameter & Type & Description \\
\midrule
$D_{\min}[d,s]$       & $\mathbb{Z}_{\geq 0}$ & Min required staffing at slot $s$ on day $d$ \\
$D_{\mathrm{ideal}}[d,s]$  & $\mathbb{Z}_{\geq 0}$ & Ideal staffing at slot $s$ on day $d$ \\
$D_{\mathrm{skill}}[k,d,s]$ & $\mathbb{Z}_{\geq 0}$ & Min employees with skill $k$ at $(d,s)$ \\
$U[e,d,s]$            & $\{0,1\}$  & 1 if employee $e$ unavailable at $(d,s)$ \\
$P[e,d,s]$            & $\{0,1\}$  & 1 if employee $e$ prefers to work at $(d,s)$ \\
$x_{\mathrm{prev}}[e,d,s]$ & $\{0,1\}$  & Previous week's schedule (for S12) \\
$L_{\min}$            & $\mathbb{R}_{>0}$  & Min daily shift length; default: 4h \\
$L_{\max}$            & $\mathbb{R}_{>0}$  & Max daily shift length; default: 10h \\
$H_{\min}$            & $\mathbb{R}_{>0}$  & Min weekly working hours; default: 20h \\
$H_{\max}$            & $\mathbb{R}_{>0}$  & Max weekly working hours; default: 48h \\
$H_{\mathrm{target,d}}$  & $\mathbb{R}_{>0}$  & Target daily hours for S6; default: 8h \\
$H_{\mathrm{target,w}}$  & $\mathbb{R}_{>0}$  & Target weekly hours for S7; default: 40h \\
$H_{\mathrm{rest}}$      & $\mathbb{R}_{>0}$  & Min rest between shifts; default: 10h \\
$C_{\max}$            & $\mathbb{Z}_{>0}$  & Max consecutive working days; default: 5 \\
$B_{\mathrm{len}}$       & $\mathbb{Z}_{>0}$  & Break duration in slots \\
$B_{\mathrm{thresh}}$    & $\mathbb{Z}_{>0}$  & Shift length triggering mandatory break \\
$k_{\mathrm{floor}}$     & $\mathbb{Z}_{>0}$  & Min employees per slot when demand exists \\
$k_{\mathrm{break}}$     & $\mathbb{Z}_{>0}$  & Max concurrent employees on break \\
$\mathit{WP}_{\mathrm{role}}[r]$ & $\mathbb{R}_{>0}$ & Workload pts/slot for role $r$ \\
$a_i$                 & $\{0,1\}$  & Activation flag for constraint $i$ \\
$w_i$                 & $\mathbb{R}$       & Weight for constraint $i$ \\
\bottomrule
\end{tabular}
\end{table}

\subsection*{B.3 Decision Variables}

A key contribution of this work is the three-variable shift-window decomposition that replaces the single binary variable $x[e,d,s] \in \{0,1\}$ used in MOGA~\cite{patel2025}. The triple $(x, w, b)$ enables mandatory break scheduling and centrality optimization, which are impossible with a single binary variable.

%  - - Table 19  - -
\begin{table}[H]
\centering
\caption{CP-SAT decision variables. $x$, $w$, $b$ are the shift-window triple; $\alpha$, $\beta$ are derived shift-boundary integer variables.}
\label{tab:decision_vars_full}
\vspace{2pt}
\scriptsize
\setlength{\tabcolsep}{2pt}
\begin{tabular}{llp{3.8cm}}
\toprule
Variable & Domain & Semantics \\
\midrule
$x[e,d,s]$ & $\{0,1\}$           & 1 iff employee $e$ actively working slot $s$ on day $d$ \\
$w[e,d,s]$ & $\{0,1\}$           & 1 iff slot $s$ within $e$'s shift window on day $d$ \\
$b[e,d,s]$ & $\{0,1\}$           & 1 iff employee $e$ on break at slot $s$ on day $d$ \\
$y[e,d]$   & $\{0,1\}$           & 1 iff employee $e$ works $\geq 1$ slot on day $d$ \\
$\alpha[e,d]$ & $\{0,\ldots,T{-}1\}$ & First active (work) slot for $e$ on day $d$ \\
$\beta[e,d]$  & $\{0,\ldots,T{-}1\}$ & Last active (work) slot for $e$ on day $d$ \\
\bottomrule
\end{tabular}
\end{table}

\subsection*{B.4 Structural Identity and Shift-Window Model}

The three binary variables $x$, $w$, $b$ are linked by the following structural identity that holds for all employees $e \in E$, days $d \in D$, slots $s \in S$:
\begin{equation}
x[e, d, s] = w[e, d, s] - b[e, d, s] \quad \forall\, e, d, s \tag{A.1}
\end{equation}

This decomposition creates three mutually exclusive states for each $(e, d, s)$ triple:

%  - - Table 20  - -
\begin{table}[H]
\centering
\caption{State space of the $(w, b, x)$ triple for each $(e, d, s)$. Only three states are feasible.}
\label{tab:state_space}
\vspace{2pt}
\scriptsize
\setlength{\tabcolsep}{3pt}
\begin{tabular}{cccl}
\toprule
$w$ & $b$ & $x$ & Interpretation \\
\midrule
0 & 0 & 0    & Off-duty \\
1 & 0 & 1    & Active work \\
1 & 1 & 0    & Scheduled break \\
0 & 1 & -   & Invalid (excluded by H10) \\
\bottomrule
\end{tabular}
\end{table}

The shift-window model enables: (1) mandatory break insertion at the CP-SAT level (H11), which is impossible with a single binary variable; (2) break centrality optimization (S11), measuring the distance between the break midpoint and the shift midpoint; and (3) concurrent break limiting (H12), preventing operational disruption from simultaneous absences. The shift start ($\alpha$) and end ($\beta$) variables are derived as:
\begin{align}
\alpha[e,d] &= \min\{s \in S : x[e,d,s] = 1\}, \notag\\
\beta[e,d] &= \max\{s \in S : x[e,d,s] = 1\}. \tag{A.2}
\end{align}

\subsection*{B.5 Objective Function}

The CP-SAT model minimizes a single weighted-sum objective $Z$ over all 15 soft penalty components. Each component can be independently activated and weighted:
\begin{equation}
\text{minimize}\; Z = \sum_{i=1}^{15} a_i \cdot w_i \cdot f_i(x, w, b, y) \tag{A.3}
\end{equation}
where $a_i \in \{0,1\}$ is the JSON-configurable activation flag and $w_i \in \mathbb{R}$ is the configurable weight (negative weight converts a penalty to a reward). The three-objective decomposition from MOGA~\cite{patel2025} is recovered by:
\begin{align}
f_1 &= \sum\nolimits_{i\in\{1,2,4,5\}} f_i, \notag\\
f_2 &= \sum\nolimits_{i\in\{3,6,7,8\}} f_i, \notag\\
f_3 &\equiv 0 \quad\text{(by construction via H1 - H14)}. \tag{A.4}
\end{align}

\subsection*{B.6 Hard Constraints (H1 - H14)}

The following 14 constraints are imposed as inviolable requirements. Any feasible CP-SAT solution satisfies all of them; violations are structurally impossible (not merely penalized).

\smallskip\noindent\textbf{H1 Empty-on-Empty:}
\begin{equation}
x[e, d, s] = 0 \quad \forall\, e\in E,\; (d,s): D_{\min}[d][s] = 0 \tag{A.5}
\end{equation}
Employees cannot be assigned to slots with zero demand.

\smallskip\noindent\textbf{H2 Unavailability:}
\begin{equation}
w[e, d, s] = 0 \quad \forall\, e\in E,\; (d,s): U[e][d][s] = 1 \tag{A.6}
\end{equation}
Shift window forbidden during declared unavailability.

\smallskip\noindent\textbf{H3 Minimum Floor Staffing:}
\begin{equation}
\sum_{e\in E} x[e, d, s] \geq k_{\mathrm{floor}} \quad \forall\, (d,s): D_{\min}[d][s] > 0 \tag{A.7}
\end{equation}

\smallskip\noindent\textbf{H4 Daily Shift Length:}
\begin{align}
& \frac{L_{\min}}{\delta} \leq \sum_{s\in S} x[e, d, s] \leq \frac{L_{\max}}{\delta}, \notag\\
& \quad \forall\, e\in E,\; d\in D: y[e,d]=1 \tag{A.8}
\end{align}

\smallskip\noindent\textbf{H5 Minimum Inter-Shift Rest:}
\begin{equation}
\alpha[e, d{+}1] - \beta[e, d] \geq \tfrac{H_{\mathrm{rest}}}{\delta} \;\; \forall\, e\in E,\; \text{consec.\ } d, d{+}1 \tag{A.9}
\end{equation}

\smallskip\noindent\textbf{H6 Maximum Consecutive Working Days:}
\begin{equation}
\sum_{j=d}^{d+C_{\max}} y[e, j] \leq C_{\max} \quad \forall\, e\in E,\; \text{valid } d\in D \tag{A.10}
\end{equation}

\smallskip\noindent\textbf{H7 Weekly Hour Limits:}
\begin{equation}
\frac{H_{\min}}{\delta} \leq \sum_{d\in D,\, s\in S} x[e, d, s] \leq \frac{H_{\max}}{\delta} \quad \forall\, e \in E_{\mathrm{real}} \tag{A.11}
\end{equation}

\smallskip\noindent\textbf{H8 Workforce Utilization:}
\begin{equation}
\sum_{d\in D} y[e, d] \geq 1 \quad \forall\, e \in E_{\mathrm{real}} \tag{A.12}
\end{equation}

\smallskip\noindent\textbf{H9 Weekly Minimum Coverage:}
\begin{equation}
\sum_{e\in E,\, d\in D,\, s\in S} x[e, d, s] \geq \sum_{d\in D,\, s\in S} D_{\min}[d][s] \tag{A.13}
\end{equation}

\smallskip\noindent\textbf{H10 Single Continuous Shift:}
\begin{align}
& w[e, d, \cdot] \text{ forms exactly one contiguous block per day,} \notag\\
& \forall\, e\in E,\; d\in D \tag{A.14}
\end{align}
Enforced via \texttt{AddBoolOr} chains that force every interior window slot ($\alpha{<}s{<}\beta$) to be active.

\smallskip\noindent\textbf{H11 Mandatory Break:}
\begin{align}
& \sum_{s\in S} w[e,d,s] \geq B_{\mathrm{thresh}} \;\Rightarrow\; \notag\\
& \quad \sum_{s\in S} b[e,d,s] = B_{\mathrm{len}} \;\wedge\; b[e,d,\cdot] \text{ contiguous} \tag{A.15}
\end{align}

\smallskip\noindent\textbf{H12 Break Concurrency Limit:}
\begin{equation}
\sum_{e\in E} b[e, d, s] \leq k_{\mathrm{break}} \quad \forall\, d\in D,\; s\in S \tag{A.16}
\end{equation}

\smallskip\noindent\textbf{H13 Weekend Management Coverage:}
\begin{align}
& \sum_{e\in E_{\mathrm{mgr}}} x[e, d, s] \geq 1, \notag\\
& \quad \forall\, d\in\{\text{Sat, Sun}\},\; s: D_{\min}[d][s]>0 \tag{A.17}
\end{align}

\smallskip\noindent\textbf{H14 Skill Coverage:}
\begin{align}
& \sum_{e:\, k\in K_e} x[e, d, s] \geq D_{\mathrm{skill}}[k][d][s], \notag\\
& \quad \forall\, \text{required skills } k,\; (d,s) \tag{A.18}
\end{align}

\subsection*{B.7 Soft Constraint Penalty Functions (S1 - S15)}

The following 15 penalty functions form the weighted-sum objective $Z$ (Eq.~A.3). S1 - S8 extend the $f_1/f_2$ objectives of MOGA~\cite{patel2025}; S9 - S15 are new quality dimensions introduced by CP-SAT. Each $f_i$ is scaled by activation flag $a_i$ and weight $w_i$. Positive $w_i$ = penalty; negative $w_i$ = reward.

\smallskip\noindent\textbf{S1 Slot Understaffing:}
\begin{equation}
f_1 = \sum_{d\in D,\, s\in S} \max\!\big(0,\; D_{\min}[d][s] - \textstyle\sum_{e\in E} x[e,d,s]\big) \tag{A.19}
\end{equation}

\smallskip\noindent\textbf{S2 Slot Overstaffing:}
\begin{equation}
f_2 = \sum_{d\in D,\, s\in S} \max\!\big(0,\; \textstyle\sum_{e\in E} x[e,d,s] - D_{\mathrm{ideal}}[d][s]\big) \tag{A.20}
\end{equation}

\smallskip\noindent\textbf{S3 Daily Understaffing:}
\begin{align}
f_3 = \sum_{d\in D} \max\big(0,\; & \textstyle\sum_{s\in S} D_{\min}[d][s] \notag\\
& {}- \textstyle\sum_{e\in E,\, s\in S} x[e,d,s]\big) \tag{A.21}
\end{align}

\smallskip\noindent\textbf{S4 Daily Overstaffing:}
\begin{align}
f_4 = \sum_{d\in D} \max\big(0,\; & \textstyle\sum_{e\in E,\, s\in S} x[e,d,s] \notag\\
& {}- \textstyle\sum_{s\in S} D_{\mathrm{ideal}}[d][s]\big) \tag{A.22}
\end{align}

\smallskip\noindent\textbf{S5 Weekly Overstaffing:}
\begin{equation}
f_5 = \max\!\big(0,\; \textstyle\sum_{e,d,s} x[e,d,s] - \sum_{d,s} D_{\mathrm{ideal}}[d][s]\big) \tag{A.23}
\end{equation}

\smallskip\noindent\textbf{S6 Daily Hours Target:}
\begin{equation}
f_6 = \sum_{\substack{e\in E,\, d\in D:\\ y[e,d]=1}} \Big|\sum_{s\in S} x[e,d,s] \cdot \delta - H_{\mathrm{target,d}}\Big| \tag{A.24}
\end{equation}

\smallskip\noindent\textbf{S7 Weekly Hours Target:}
\begin{equation}
f_7 = \sum_{e\in E} \Big|\sum_{d\in D,\, s\in S} x[e,d,s] \cdot \delta - H_{\mathrm{target,w}}\Big| \tag{A.25}
\end{equation}

\smallskip\noindent\textbf{S8 Missing Management Coverage:}
\begin{align}
f_8 = \sum_{d,s} \max\Big(0,\; & \mathbb{1}\big(\textstyle\sum_e x[e,d,s]>0\big) \notag\\
& {}- \textstyle\sum_{e\in E_{\mathrm{mgr}}} x[e,d,s]\Big) \tag{A.26}
\end{align}

\smallskip\noindent\textbf{S9 Management Overlap:}
\begin{equation}
f_9 = \sum_{d\in D,\, s\in S} \max\!\big(0,\; \textstyle\sum_{e\in E_{\mathrm{mgr}}} x[e,d,s] - 1\big) \tag{A.27}
\end{equation}

\smallskip\noindent\textbf{S10 Management Open/Close Reward:}
\begin{equation}
f_{10} = -\sum_{d\in D} \sum_{e\in E_{\mathrm{mgr}}} \big(x[e,d,s_{\mathrm{open}}] + x[e,d,s_{\mathrm{close}}]\big) \tag{A.28}
\end{equation}

\smallskip\noindent\textbf{S11 Break Centrality:}
\begin{equation}
f_{11} = \sum_{e,d:\, b>0} |\mathrm{mid}(b[e,d,\cdot]) - \mathrm{mid}(w[e,d,\cdot])| \tag{A.29}
\end{equation}

\smallskip\noindent\textbf{S12 Inter-Week Stability:}
\begin{equation}
f_{12} = \sum_{e\in E,\, d\in D,\, s\in S} \big|x[e,d,s] - x_{\mathrm{prev}}[e,d,s]\big| \tag{A.30}
\end{equation}

\smallskip\noindent\textbf{S13 Intra-Week Stability:}
\begin{align}
f_{13} = \sum_{e\in E} \sum_{\substack{d_1<d_2 \\ d_1,d_2\in D}} \big( & |\alpha[e,d_1]{-}\alpha[e,d_2]| \notag\\
& + |\beta[e,d_1]{-}\beta[e,d_2]|\big) \tag{A.31}
\end{align}

\smallskip\noindent\textbf{S14 Preferred Hours Reward:}
\begin{equation}
f_{14} = -\sum_{e\in E,\, d\in D,\, s\in S} P[e,d,s] \cdot x[e,d,s] \tag{A.32}
\end{equation}

\smallskip\noindent\textbf{S15 Workload Equity / Min-Max Fairness:}
\begin{equation}
f_{15} = \max_{e\in E_{\mathrm{real}}} \big|\mathit{WP}(e) - \mathit{WP}_{\mathrm{baseline}}(e)\big| \tag{A.33}
\end{equation}
Min-max fairness: minimizes the maximum deviation from each employee's role-based workload baseline. $\mathit{WP}(e) = \sum_{d,s} x[e,d,s] \cdot (\mathit{WP}_{\mathrm{role}}[\mathrm{role}(e)] + \mathit{skill\_bonus}[e,d,s] + \mathit{intensity}[d,s])$.

\subsection*{B.8 Workload Point Computation (for S15)}

The Workload Point (WP) score for employee $e$ integrates three components that capture both the volume and difficulty of assignments:
\begin{multline}
\mathit{WP}(e) = \sum_{d\in D,\, s\in S} x[e,d,s] \cdot \big(\mathit{WP}_{\mathrm{role}}[\mathrm{role}(e)] \\
 + \mathit{skill\_bonus}[e,d,s] + \mathit{intensity}[d,s]\big) \tag{A.34}
\end{multline}

The three components are:
\begin{itemize}
\item $\mathit{WP}_{\mathrm{role}}[r]$: Base load by role. L1/L2 (supervisors): 120 pts/slot; L3: 100 pts/slot; L4: 90 pts/slot.
\item $\mathit{skill\_bonus}[e,d,s]$: +15 pts when employee $e$ uses a critical skill at $(d,s)$; 0 otherwise.
\item $\mathit{intensity}[d,s]$: Bidirectional slot-pressure factor. Understaffed slots ($D_{\min} >$ current coverage) receive positive intensity; overstaffed slots receive negative intensity. This makes high-demand coverage inherently more workload-heavy, capturing true operational burden.
\end{itemize}

The equity objective $f_{15} = \max_{e} |\mathit{WP}(e) - \mathit{WP}_{\mathrm{baseline}}(e)|$ minimizes the worst-case deviation from each employee's expected workload, ensuring no single employee bears a disproportionate burden relative to their~role.

% ============================================================
% APPENDIX C
% ============================================================
\section*{Appendix C: Configuration Reference}

All scheduling rules are externalised to \texttt{config.json}:

{\small
\begin{verbatim}
{
  "store_id": "string",
  "solver_run_time": 600,
  "Operational_Rules": {
    "Scheduling_Granularity": "30_MIN",
    "Max_daily_work_hour_limit": 10,
    "Min_daily_work_hour_limit": 3,
    "Max_weekly_work_hour_limit": 60,
    ...
  },
  "Constraint_Weights": {
    "slot_understaffing": 1.0,
    "workload_equity": 10.0,
    "break_centrality": 10.0,
    ...
  },
  "Constraint_Activation": {
    "check_mandatory_break": true,
    "check_workload_equity": true,
    "check_intra_week_stability": true,
    ...
  },
  "Staff_Roles": { ... },
  "Staff_Skills": { ... },
  "Skill_Requirements": { ... }
}
\end{verbatim}
}

\texttt{Constraint\_Activation} contains one boolean flag per constraint. \texttt{Constraint\_Weights} contains one float weight per soft constraint. Both can be modified at runtime via the agent's \texttt{update\_config\_tool} without restarting the application.

\subsection*{C.1 CP-SAT Solver Configuration}

CP-SAT uses Google OR-Tools v9.12~\cite{perron2024} with the configuration in Table~\ref{tab:solver_config}. The portfolio combines 11 full-problem workers, 5 first-solution workers and 11 LNS workers (RINS, RENS, graph-based, random), providing robust performance without manual algorithm tuning. All 29 constraints are independently activatable via JSON flags with no code changes required.

%  - - Table 21  - -
\begin{table}[H]
\centering
\caption{CP-SAT solver configuration.}
\label{tab:solver_config}
\vspace{2pt}
\footnotesize
\setlength{\tabcolsep}{3pt}
\begin{tabular}{llp{3.5cm}}
\toprule
Parameter & Value & Description \\
\midrule
Search Workers   & 16       & Parallel search threads with diverse strategies \\
Symmetry Level   & 3        & Maximum symmetry detection and breaking \\
Time Limit       & 120\,s / unit & Wall-clock ceiling per instance \\
LNS Workers      & 11       & RINS, RENS, graph-based, random LNS subsolvers \\
Presolve         & Full     & Propagation, symmetry, dual reasoning (${\sim}60$-$70\%$ size reduction) \\
\bottomrule
\end{tabular}
\end{table}

\end{document}